\documentclass[11pt, a4paper, logo, copyright, nonumbering]{xiaomi}

\usepackage[authoryear, sort&compress, round]{natbib}
\usepackage{dblfloatfix}
\usepackage{ulem}
\usepackage{caption}
\usepackage{dramatist}
\usepackage{xspace}
\usepackage{pifont} 
\usepackage{multirow}
\usepackage{tcolorbox}
\usepackage{xltabular}
\usepackage{longtable}
\usepackage{hyperref}
\usepackage{wrapfig}

\usepackage{amsfonts}
\usepackage{amsmath}
\usepackage{amssymb}
\usepackage{lineno}
\usepackage{multirow}
\usepackage{adjustbox}

\usepackage[bottom]{footmisc}

\usepackage{CJKutf8}
\usepackage{setspace}
\usepackage{makecell}
\usepackage{graphicx}
\usepackage{subcaption}
\usepackage{multicol} 

\newcommand{\mimo}{MiMo-7B}

\newcommand{\mimoflash}{MiMo-V2-Flash}

\definecolor{xiaomiorange}{HTML}{FF6901}

\title{\centering \mimoflash{} Technical Report}

\author{
 LLM-Core Xiaomi
}

\begin{abstract}

We present \mimoflash{}, a Mixture-of-Experts (MoE) model with 309B total parameters and 15B active parameters, designed for fast, strong reasoning and agentic capabilities. 
\mimoflash{} adopts a hybrid attention architecture that interleaves \textbf{Sliding Window Attention (SWA)} with global attention, with a 128-token sliding window under a 5:1 hybrid ratio. 
The model is pre-trained on 27 trillion tokens with \textbf{Multi-Token Prediction (MTP)}, employing a native 32k context length and subsequently extended to 256k.
To efficiently scale post-training compute, \mimoflash{} introduces a novel \textbf{Multi-Teacher On-Policy Distillation (MOPD)} paradigm. 
In this framework, domain-specialized teachers (e.g., trained via large-scale reinforcement learning) provide dense and token-level reward, enabling the student model to perfectly master teacher expertise. 
MiMo-V2-Flash rivals top-tier open-weight models such as DeepSeek-V3.2 and Kimi-K2, despite using only $1/2$ and $1/3$ of their total parameters, respectively.
During inference, by repurposing MTP as a draft model for speculative decoding, \mimoflash{} achieves up to 3.6 acceptance length and 2.6$\times$ decoding speedup with three MTP layers.
We open-source both the model weights and the three-layer MTP weights to foster open research and community collaboration.

\end{abstract}

\begin{document}

\maketitle

\begin{figure*}[h] 
    \centering
    \includegraphics[width=1\linewidth]{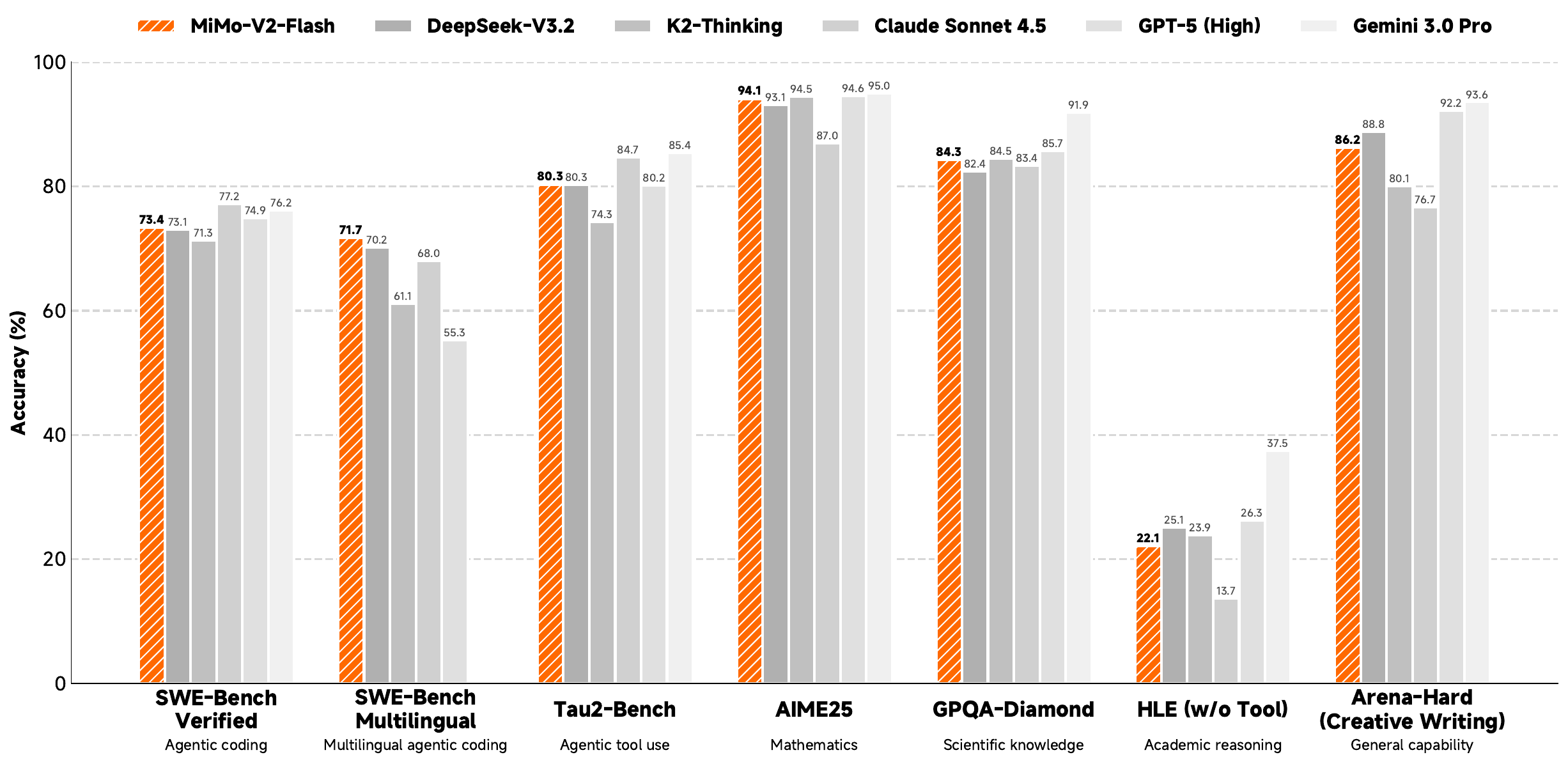}
    \caption{Benchmark performance of MiMo-V2-Flash.}
    \label{fig:teaser}
\end{figure*}

\newpage

\begin{spacing}{0.9}
\tableofcontents
\end{spacing}

\newpage

\section{Introduction}

Recent progress towards Artificial General Intelligence (AGI) is increasingly propelled by two frontiers: advanced reasoning chains and autonomous agentic workflows~\citep{gemini3pro,liu2025deepseek,team2025kimi}, grounded in large-scale Reinforcement Learning (RL).
Yet building scalable reasoners and agents hits a common critical bottleneck, where long-context modeling must be simultaneously fast and strong.

In this work, we introduce \mimoflash{}, an efficient and cost-effective Large Language Model (LLM) that delivers strong reasoning and agentic performance.
\mimoflash{} is a 309B-parameter MoE with 15B activated per token. 
To alleviate the quadratic complexity of full attention, \mimoflash{} adopts a hybrid attention mechanism that interleaves local sliding window and global attention. The sliding window size is 128-token and the hybrid local:global ratio is 5:1, yielding nearly a 6× reduction in KV-cache storage and attention computation for long contexts.
With the help of learnable attention sink bias~\citep{agarwal2025gpt}, the hybrid architecture maintains strong modeling capability even in long-context scenarios, despite the aggressive sliding window size and hybrid ratio.
\mimoflash{} also incorporates Multi-Token Prediction (MTP) to enhance training performance and accelerate inference decoding. In particular, MTP has strong potential to boost RL rollout speed, which helps to scale LLMs towards greater intelligence. 
With a lightweight dense Feed-Forward Network (FFN) and sliding window attention, our MTP block delivers substantial decoding speedups in practice at high acceptance rates.

The pre-training recipe of \mimoflash{} largely follows that of \mimo{}~\citep{xia2025mimo}, with several enhancements. Training is conducted using FP8 mixed-precision, enabling efficient large-scale training over 27T tokens. The model is initially pre-trained with a native 32K context and later extended to 256K. The resulting pretrained model, \mimoflash{}-Base, has been evaluated against leading open-source base models such as Kimi-K2-Base~\citep{team2025kimik2} and DeepSeek-V3.2-Exp-Base~\citep{liu2025deepseek}. 
MiMo-V2-Flash-Base achieves competitive performance across general benchmarks and surpasses peer models on reasoning-focused tasks.
For long-context retrieval, our hybrid attention architecture achieves nearly 100\% success rates across context lengths from 32K to 256K. On the extreme long-context reasoning benchmark GSM-Infinite~\citep{zhou2025gsm}, MiMo-V2-Flash demonstrates robust performance with minimal degradation when scaling from 16K to 128K.

In post-training, we focus on efficiently scaling RL compute to improve reasoning and agentic capabilities. To this end, \mimoflash{} introduces a novel post-training paradigm termed Multi-Teacher On-Policy Distillation (MOPD). This framework addresses both learning inefficiency and capability imbalance through a three-stage process: (1) general Supervised Fine-Tuning (SFT); (2) specialized RL/SFT to train domain-specific teacher models; (3) MOPD, wherein the student model learns from two complementary signals: dense, token-level rewards from specialized teachers trained across diverse domains, and a verifiable, outcome-based reward. By integrating diverse expert knowledge in this manner, \mimoflash{} simultaneously masters the peak capabilities of domain teachers while benefiting from stable and efficient learning dynamics.

\mimoflash{} achieves performance comparable
to that of Kimi-K2-Thinking and DeepSeek-V3.2-Thinking on most reasoning benchmarks.
In long-context evaluations such as LongBench V2 and MRCR, \mimoflash{} consistently surpasses larger full-attention models, confirming the robustness of its hybrid SWA architecture. Notably, the model attains 73.4\% on SWE-Bench Verified and 71.7\% on SWE-Bench Multilingual, establishing it as the leading open-source model for software engineering tasks.
The model weights (with 3-layer MTP weights) are available at 
\textcolor{xiaomiorange}{\url{https://github.com/XiaomiMiMo/MiMo-V2-Flash}}.

\section{\mimoflash{} Model Architecture}

\subsection{Overall Architecture}
\begin{figure}[t!]
    \centering
    \includegraphics[width=0.8\linewidth]{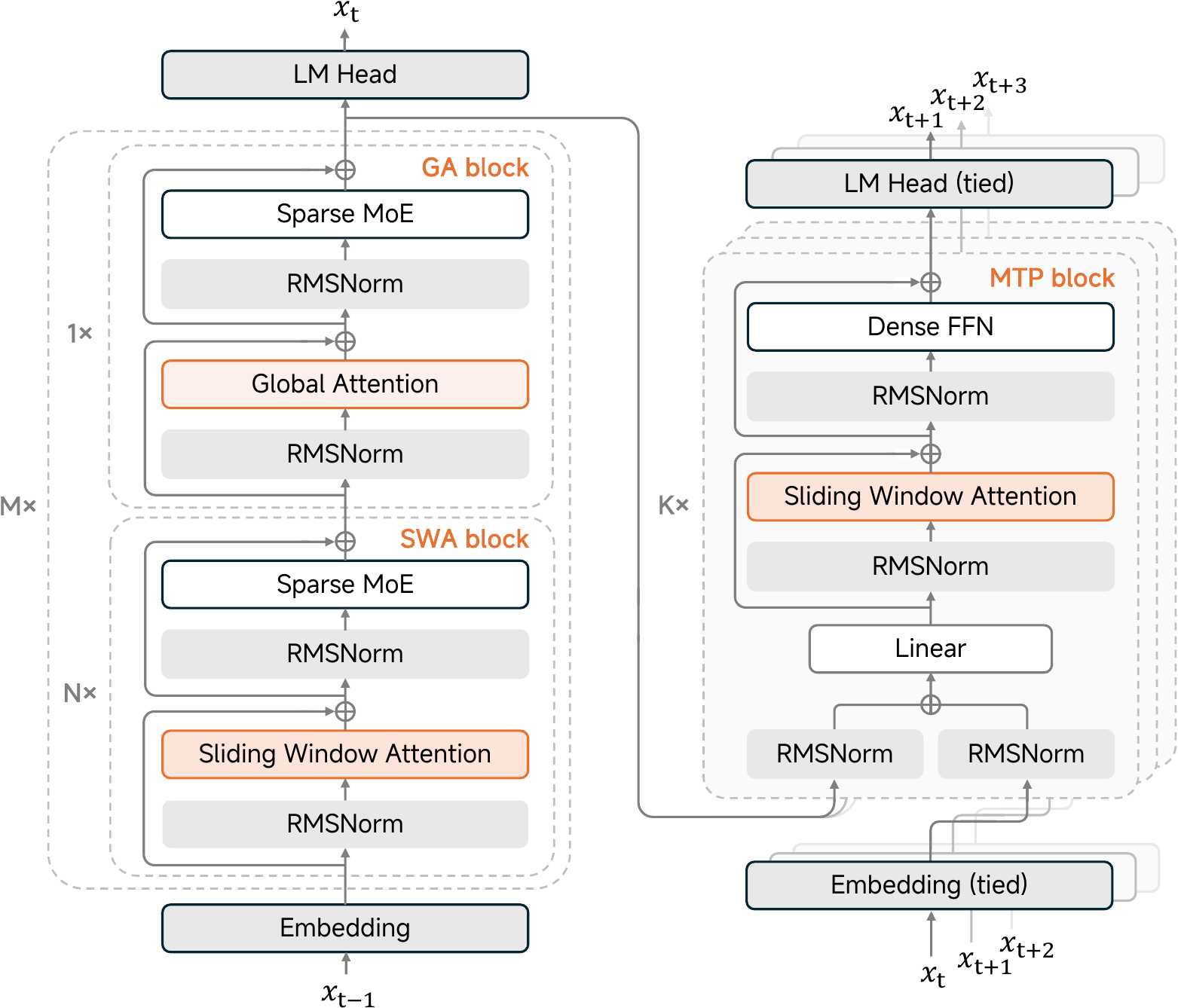}
    \caption{An illustration of \mimoflash{} model architecture. The model comprises $M=8$ Hybrid Blocks, where each Hybrid Block interleaves $N=5$ Sliding Window Attention (SWA) blocks with one Global Attention (GA) block. Both are equipped with a sparse MoE FFN. The only exception is the first block, which uses GA with a dense FFN. The MTP blocks employ SWA and a dense FFN.}
    \label{fig:model_arch}
\end{figure}

As illustrated in Figure~\ref{fig:model_arch}, \mimoflash{} follows a standard Transformer~\citep{vaswani2017attention} backbone augmented with MoE~\citep{shazeer2017outrageously} and hybrid attention~\citep{gpt3, gemma2, team2025gemma, li2025minimax, qwen3next2025, kda}. 
\mimoflash{} is mainly composed of repeated hybrid blocks that interleave Local Sliding Window Attention (SWA) and Global Attention (GA).
It stacks $M = 8$ hybrid blocks, each structured with $N = 5$ consecutive SWA blocks followed by an GA block. 
The only exception is the very first Transformer block, which uses global attention with a dense Feed-Forward Network (FFN) to stabilize early representation learning. The sliding window size $W$ used in \mimoflash{} is $128$. Both the SWA block and the GA block utilize the sparse MoE FFN. 
Each MoE layer comprises 256 experts in total, with 8 activated per token, and contains no shared experts.

\mimoflash{} also integrates MTP~\citep{gloeckle2024better, liu2024deepseek, xia2025mimo} to improve model performance (both quality and efficiency). Worth noting, the MTP block uses dense FFN instead of MoE and applies SWA rather than GA, making it lightweight for speculative decoding. The number of parameters for each MTP block is only 0.33B.

\begin{table}[ht]
\small
\centering
\begin{tabular}{llc}
\toprule
\textbf{Block} & \textbf{Configuration} & \textbf{Value} \\
\midrule

\multirow{7}{*}{\textbf{Main Block}}
& Layers (Total/SWA/GA)           & $48 / 39 / 9$ \\
& SWA Heads (Q/KV)                & $64 / 8$ \\
& Sliding Window Size    & $128$ \\
& GA Heads (Q/KV)                 & $64 / 4$ \\
& Head Dimensions (QK/V)          & $192 / 128$ \\
& Experts (Total/Activated)       & $256 / 8$ \\
\midrule

\multirow{4}{*}{\textbf{MTP Block}}
& SWA Heads (Q/KV)                & $64 / 8$ \\
& Sliding Window Size    & $128$ \\
& Head Dimensions (QK/V)          & $192 / 128$ \\
& \# Parameters            & $0.33$B \\

\bottomrule
\end{tabular}
\caption{Detailed model configuration of MiMo-V2-Flash.}
\label{tab:model_config}
\end{table}

Table \ref{tab:model_config} summarizes detailed configurations of \mimoflash{}. The model consists of 39 SWA layers and 9 GA layers. Both SWA and GA utilize Grouped-Query Attention (GQA)~\citep{ainslie2023gqa}. Specifically,
SWA has 64 query heads and 8 key-value heads, while GA has 64 query heads and 4 key-value heads. The per-head dimensions are the same for SWA and GA (192 for queries and keys, and 128 for values). Rotary Positional Embedding (RoPE, ~\cite{su2024roformer}) is partially applied to the first 64 dimensions query and key. 
Following recent best practices, we adopt an FP8 mixed-precision framework similar to DeepSeek-V3~\citep{liu2024deepseek}.
Specifically, we retain BF16 precision for the attention output projections, as well as for the embedding and output head parameters, while maintaining FP32 precision for the MoE router parameters.
This mixed-precision configuration improves numerical stability without materially impacting training efficiency or memory footprint.

\subsection{Hybrid Sliding Window Attention Architecture}

Sliding window attention~\citep{beltagy2020longformer} restricts each token’s attention scope to a local window rather than the entire sequence, thereby reducing both computational and memory complexity dramatically.
This naturally motivates hybrid attention architectures that interleave sliding window attention with global attention.
However, prior work has shown that overly aggressive use of SWA, such as very small sliding window sizes or high SWA:GA ratios, can lead to substantial degradation in model performance~\citep{team2025gemma}, especially in long-context tasks.
Recently, the introduction of learnable attention sink bias, which allows the model to assign little or no attention to tokens when needed, has substantially enhanced the modeling capacity of SWA-based architectures~\citep{agarwal2025gpt}.
While the precise theoretical underpinnings of the attention sink mechanism remain an active research area~\citep{xiao2023efficient, attnsink_study, qiu2025gatedattn, sun2024massive}, we empirically observed that learnable attention sinks bias dramatically enhance the performance of hybrid SWA models, matching or even surpassing baselines with fully GA layers.

In \mimoflash{}, our implementation follows the design used in gpt-oss~\citep{agarwal2025gpt}, where a learnable attention sink bias $sink \in \mathbb{R}$ is applied to the denominator of softmax for each attention head. Specifically, let the attention logits between token $i$ and $j$ of one single head be:
\begin{equation}
a_{ij} = \frac{q_ik_j^\top}{\sqrt{d}},
\end{equation}
where $q_i$ and $k_j$ denote the query of token $i$ and key of token $j$, respectively, and $d$ is the head dimension.  
The attention weights are then given by:
\begin{equation}
s_{ij}
= \frac{\exp \left( a_{ij} - m_i \right)}
{\exp \left( sink - m_i \right) + \sum_{j'} \exp \left( a_{ij'} - m_i \right)},
\end{equation}

\begin{equation}
m_i = \max \left( \max_j a_{ij}, \; sink \right).
\end{equation}

Finally, the attention output for query $i$ is obtained as a weighted sum over the values:
\begin{equation}
o_i = \sum_{j=1}^{n} s_{ij} v_j.
\end{equation}

\subsubsection{Model Architecture Experiments}
To validate the effectiveness of our design choice, we conduct exploratory and empirical studies on a 32B dense model, maintaining the query–key dimensions and rotary embedding configurations consistent with those described above.

\begin{table}[h]
\centering
\small
\begin{tabular}{lccccccc}
\toprule
\textbf{Model} & MMLU & BBH & TriviaQA & GSM8K & MATH & CMMLU & MBPP \\
\midrule
All GA          & 57.3 & 54.7 & 53.2 & 34.2 & 9.5 & 50.3 & 54.7  \\
Hybrid SWA($W = 128$, w/o sink)  & 54.9 & 52.4 & 52.8 & 36.9 & 8.9 & - & - \\
Hybrid SWA($W = 128$, w/ sink)    & \textbf{58.3} & \textbf{56.1} & 53.7 & 36.9 & \textbf{10.3} & \textbf{53.3} & \textbf{56.3} \\
Hybrid SWA($W = 512$, w/ sink)    & \textbf{58.3} & 54.9 & \textbf{54.9} & \textbf{37.9} & 10.0 & 52.3 & 53.2 \\
\bottomrule
\end{tabular}
\caption{General benchmark results for different attention configurations.}
\label{tab:general_ablation}
\end{table}

\begin{table}[h]
\centering
\small
\begin{tabular}{lcccc}
\toprule
\textbf{Model} & GSM-Infinite & NoLiMa & RULER-32k & MRCR  \\
\midrule
All GA          & 12.3 & 49.7 & \textbf{89.4} & 32.5   \\
Hybrid SWA($W = 128$, w/ sink)           & \textbf{17.3} & \textbf{51.2} & \textbf{89.4} & \textbf{34.4}   \\
Hybrid SWA($W = 512$, w/ sink)    & 17.2 & 38.5 & 84.7 & 19.6   \\
\bottomrule
\end{tabular}
\caption{Long-context benchmark results for different attention configurations.}
\label{tab:longcontext_ablation}
\end{table}

\begin{table}[h!]
\centering
\small
\begin{tabular}{lcccc}
\toprule
\textbf{Model} & AIME24/25 & LiveCodebench & GPQA-Diamond & Average \\
\midrule
All GA          & 45.5 & 40.0 & 41.7 & 42.4  \\
Hybrid SWA($W = 128$, w/ sink)           & \textbf{47.1} & \textbf{43.9} & \textbf{48.1} & \textbf{46.3}   \\
\bottomrule
\end{tabular}
\caption{Complex reasoning benchmark results for different attention configurations.}
\label{tab:reasoning_ablation}
\end{table}

\paragraph{Baselines and Benchmarks} We evaluate four model architecture variants in a comparative setting. These include an all global attention (All GA) baseline, a hybrid SWA model with a 128-token window without attention sinks bias, and two hybrid SWA models augmented with attention sinks bias using window sizes of 128 and 512, respectively.
All variants share the same training pipeline: pre-training on 250B tokens with an 8,192 sequence length, long context extension to 32,768 over an additional 40B tokens, followed by long-context SFT and reasoning SFT with chain-of-thought supervision.
We evaluate model variants across benchmarks covering general capability, long-context understanding, and complex reasoning. General-domain results (Table~\ref{tab:general_ablation}) are obtained from pre-trained base models without long-context extension, evaluating general knowledge and reasoning on MMLU~\cite{hendrycks2020measuring}, BBH~\citep{suzgun2022challenging}, TriviaQA~\citep{joshi2017triviaqa}, GSM8K~\citep{cobbe2021training},  MATH~\citep{hendrycks2021measuring}, CMMLU~\citep{li2023cmmlu}, and MBPP~\citep{austin2021program}. Long-context results (Table~\ref{tab:longcontext_ablation}) evaluate long-context–extended base models on GSM-Infinite~\citep{zhou2025gsm}, NoLiMa~\citep{modarressi2025nolima}, and RULER-32k~\citep{hsieh2024ruler}, and long-context SFT models on MRCR~\citep{vodrahalli2024michelangelo}. For GSM-Infinite and Nolima, we construct internal few-shot benchmarks to assess base models under controlled long-context settings. Complex reasoning results (Table~\ref{tab:reasoning_ablation}) evaluate the reasoning SFT models on AIME24\&25~\citep{AIME}, LiveCodeBench~\citep{jain2024livecodebench}, and GPQA-Diamond~\citep{rein2024gpqa}. 

We highlight our key empirical findings below:

\paragraph{Ablation on Attention Sink Bias} As shown in Table~\ref{tab:general_ablation}, hybrid SWA ($W = 128$, w/o sink) suffers noticeable performance degradation across general benchmarks, whereas introducing attention sink bias consistently recovers or improves performance relative to the all-GA baseline. Thus, in our further experiments, we assume that the attention sink bias is applied by default. 

\paragraph{Sliding Window Attention Size} Hybrid SWA ($W = 128$) and Hybrid SWA ($W = 512$) appear to perform similarly on general benchmarks (Table~\ref{tab:general_ablation}). However, after long-context extension and long-context SFT, hybrid SWA ($W = 128$) surpasses the all-GA baseline, whereas SWA ($W = 512$) experiences significant degradation (Table~\ref{tab:longcontext_ablation}). 

\paragraph{Reasoning Ability} As shown in Table~\ref{tab:reasoning_ablation}, hybrid SWA ($W = 128$) surpasses the all-GA baseline across different challenging reasoning benchmarks, showing clear improvements on complex reasoning abilities.

\subsubsection{Summary and Discussion}
Our experiments show that hybrid SWA ($W = 128$) not only outperforms hybrid SWA ($W = 512$) but can also surpass the all-GA baseline, which may seem counterintuitive. We hypothesize that this arises from a combination of better regularization and effective sparsity. Smaller windows force the model to focus on local context, serving as an inductive bias that mitigates overfitting on spurious patterns. Moreover, a tighter window ($W = 128$) compels SWA to model local information while delegating long-range dependencies to the global attention layers, resulting in a clearer division of labor with more accurate and efficient learning. In contrast, a larger window ($W = 512$) can blur this distinction, causing SWA to partially handle long-range dependencies itself, which dilutes the separation between local and global information and leads to suboptimal performance.

We emphasize that these observations and findings are empirical and derived from our specific experimental settings, including model scale, datasets, and training procedures.
Nonetheless, we hope these observations contribute an additional perspective to the ongoing discussion of efficient attention architectures in the era of reasoning and agentic AI models, and motivate further community-wide investigation into efficient architecture.

\subsection{Lightweight Multi-Token Prediction (MTP)}
\subsubsection{Motivation of using MTP}
Prior work demonstrates that MTP is a powerful training objective that enhances training efficiency and model quality~\citep{gloeckle2024better,liu2024deepseek,xia2025mimo}.
Beyond these training benefits, we place stronger emphasis on exploiting MTP as a native draft model for self-speculative decoding to deliver real-deployment speedup. 
In the following, we elaborate on how MTP accelerates inference from two perspectives: general LLM decoding speedup and RL training acceleration.

\paragraph{Accelerating LLM Decoding}
LLM decoding is inherently memory-bound due to low arithmetic intensity. Batch-level parallelism is commonly used to increase FFN arithmetic intensity but does not benefit attention computation, as each request maintains its own KV cache. In contrast, MTP lifts the arithmetic intensity of both FFN and attention by generating multiple draft tokens, which the main model then verifies in parallel. This approach enables token-level parallelism without increasing KV cache I/O.

\paragraph{Accelerating RL Training}
MTP acceleration is particularly well-suited for RL training~\citep{miles2025}, where the rollout phase consistently emerges as the dominant bottleneck due to the inference and decoding costs.
MTP addresses two key challenges in RL training:
\begin{itemize}
    \item 
    \textit{It enables efficient and effective RL with small batches.}
    Current RL training relies on large-batch, off-policy algorithms to maximize throughput~\citep{schulman2017proximal,zheng2025group,liu2025deepseek}.
    However, on-policy training is generally more stable and effective, yet its small batches underutilize GPU resources.
    MTP mitigates this limitation by scaling token-level parallelism instead of batch size, making small-batch, on-policy RL training more practical.
    \item 
    \textit{It mitigates GPU idleness from long-tail stragglers.} 
    As the rollout phase progresses, long-tail stragglers that process long sequences with small batch sizes (often approaching 1) can cause significant GPU idleness~\citep{zhong2025optimizing,gao2025rollpacker}. 
    In such scenarios, MTP enhances the computational efficiency of both attention and FFN, substantially reducing overall latency.
\end{itemize}

\subsubsection{Lightweight MTP Design in \mimoflash{}}
In \mimoflash{}, the MTP block is deliberately kept lightweight to prevent it from becoming a new inference bottleneck. We use a small dense FFN rather than MoE to limit parameter count, and employ SWA instead of Global Attention (GA) to reduce KV cache and attention computation costs.
During pre-training, only a single MTP head is attached to the model to avoid extra training overhead.
In post-training, this head is replicated $K$ times to form a $K$-step MTP module, and all heads are jointly trained for multi-step prediction.
Each head receives the main-model hidden state and token embedding as input, providing richer predictive information.
Despite its lightweight design, the MTP module remains highly effective and achieves a high acceptance rate.
Detailed results are presented in Section~\ref{sec:mtp}.

\section{Pre-Training}

The \mimoflash{} pre-training corpus consists of 27 trillion tokens drawn from a diverse collection of high-quality sources, including public web content, books, academic papers, code, mathematics, and broader STEM materials. Our data processing pipeline largely follows that of MiMo-7B~\citep{xia2025mimo}, with a deliberate shift toward data exhibiting long-range dependencies.
In particular, we emphasize long-form web documents and carefully curated code corpora such as repository-level code, pull requests, issues, and commit histories to strengthen the model’s ability to capture extended contextual relationships and perform complex, multi-step reasoning.

\subsection{Data Scheduler} The pre-training of \mimoflash{} is organized into three sequential stages:
\begin{itemize}

\item \textit{Stage 1 (Pre-training, 0 -- 22T).} The model is trained on a diverse, high-quality general-purpose corpus using a context length of 32K tokens to establish strong foundational language capabilities.

\item \textit{Stage 2 (Mid-training, 22 -- 26T).} We modify the data mixture by upsampling code-centric data and incorporating approximately 5\% synthetic reasoning data to further enhance logical reasoning and program synthesis abilities.

\item \textit{Stage 3 (Context Extension, 26 -- 27T).} Following the Stage 2 data distribution, we extend the model’s context window to 256K tokens and upsample data with long-range dependencies, enabling more effective modeling of extended contexts and long-horizon reasoning.

\end{itemize}

\subsection{Hyper-Parameters}
\paragraph{Model Hyper-Parameters} We configure \mimoflash{} with 48 Transformer layers, comprising 39 sliding window attention layers and 9 global attention layers. The hidden dimension is set to 4096. All layers except the first are equipped with sparse MoE. Each MoE layer contains 256 routed experts, with 8 experts activated per token, and an intermediate hidden dimension of 2048 for each expert. The intermediate hidden dimension of the FFN of dense layers is set to 16384. All learnable parameters are randomly initialized with a standard deviation of 0.006. The model uses a single MTP layer during pre-training. Overall, \mimoflash{} has 309B total parameters, of which 15B are active.

\paragraph{Training Hyper-Parameters} 
We employ the AdamW optimizer with $\beta_1=0.9$, $\beta_2=0.95$, and a weight decay of 0.1. 
Gradient clipping is applied with a maximum norm of 1.0. 
The learning rate schedule operates in two stages.
In Stage 1, the learning rate starts with a linear warm-up from 0 to $3.2 \times 10^{-4}$ over the first 50B tokens, followed by a constant phase at $3.2 \times 10^{-4}$ for 12T tokens, and concludes with a cosine decay to $1.0 \times 10^{-4}$ over 10T tokens. Stage 2 begins at $1.0 \times 10^{-4}$ and follows a cosine decay down to $3.0 \times 10^{-5}$ over 4T tokens. The batch size warms up linearly to 2,048 over the initial 500B tokens and remains constant for the remainder of both stages.
Regarding auxiliary losses, the MoE sequence auxiliary loss coefficient is set to $1.0 \times 10^{-5}$ for all stages. The expert bias update factor is set to 0.001 during Stage 1 and Stage 2. The MTP loss weight is set to 0.3 for Stage 1 and 0.1 for Stage 2 and 3.

\paragraph{Long Context Extension} 

In Stage 1, we set the pre-training sequence length to 32,768 with a RoPE base frequency of 640,000 for GA and 10,000 for SWA. In Stage 3, the sequence length is extended to 262,144, and the RoPE base frequency of GA is adjusted to 5,000,000. The learning rate in Stage 3 decays from $3.0 \times 10^{-5}$ to $1.0 \times 10^{-5}$ following a cosine schedule, with a fixed batch size of 256. The expert bias update factor is reduced to $1.0 \times 10^{-5}$ in Stage 3.

\subsection{Evaluations}
\subsubsection{Evaluation Setup}
We evaluate \mimoflash{}-Base on a series of benchmarks, encompassing various capabilities: (1) General language understanding and reasoning, including BBH~\citep{suzgun2022challenging}, MMLU~\citep{hendrycks2020measuring}, MMLU-Redux~\citep{gema2024we}, MMLU-Pro~\citep{wang2024mmlu}, DROP~\citep{dua2019drop}, ARC~\citep{clark2018think}, HellaSwag~\citep{zellers2019hellaswag}, WinoGrande~\citep{sakaguchi2021winogrande}, TriviaQA~\citep{joshi2017triviaqa}, GPQA-Diamond~\citep{rein2024gpqa}, SuperGPQA~\citep{du2025supergpqa}, and SimpleQA~\citep{simpleqa}.
(2) Mathematics reasoning.
GSM8K~\citep{cobbe2021training}, MATH~\citep{hendrycks2021measuring}, and  AIME~\citep{AIME} (2024 \& 2025). 
(3) Coding. HumanEval+~\citep{liu2023your}, MBPP+~\citep{liu2023your}, CRUXEval~\citep{gu2024cruxeval}, MultiPL-E~\citep{cassano2022multipl}, BigCodeBench~\citep{zhuo2024bigcodebench}, LiveCodeBench-v6~\citep{jain2024livecodebench}, and SWE-Bench~\citep{jimenez2023swe} (few-shot Agentless Repair~\citep{xia2024agentless}).
(4) Chinese understanding. C-Eval~\citep{huang2023c}, CMMLU~\citep{li2023cmmlu}, and C-SimpleQA~\citep{he2025chinese}.
(5) Multilingual understanding. GlobalMMLU~\citep{singh2025global}, and INCLUDE~\citep{romanou2024include}.
(6) Long context. NIAH-Multi~\citep{hsieh2024ruler}, GSM-Infinite~\citep{zhou2025gsm} (5-shot, Hard Ops-\{2,4,6,8,10\}).

\subsubsection{Evaluation Results}

\begin{table}[htbp]
    \centering
    \small
    \resizebox{0.99\textwidth}{!}{
    \begin{tabular}{l c | c | c c c}
    \toprule
    \multirow{2}{*}{\textbf{Benchmark}} & \multirow{2}{*}{\textbf{\# Shots}} & \textbf{MiMo-V2-Flash} & \textbf{Kimi-K2} & \textbf{DeepSeek-V3.1} & \textbf{\textbf{DeepSeek-V3.2}} \\
    & & \textbf{Base} & \textbf{Base} & \textbf{Base} & \textbf{Exp Base} \\
    \midrule
    \#Activated Params & - & 15B & 32B & 37B & 37B \\
    \#Total Params & - & 309B & 1043B & 671B & 671B \\
    \midrule
    \multicolumn{6}{l}{\textbf{General}} \\
    BBH & 3-shot & 88.5 & \textbf{88.7} & 88.2 & \textbf{88.7} \\
    MMLU & 5-shot & 86.7 & \textbf{87.8} & 87.4 & \textbf{87.8} \\
    MMLU-Redux & 5-shot & \textbf{90.6} & 90.2 & 90.0 & 90.4 \\
    MMLU-Pro & 5-shot & \textbf{73.2} & 69.2 & 58.8 & 62.1 \\
    DROP & 3-shot & 84.7 & 83.6 & 86.3 & \textbf{86.6} \\
    ARC-Challenge & 25-shot & 95.9 & \textbf{96.2} & 95.6 & 95.5 \\
    HellaSwag & 10-shot & 88.5 & \textbf{94.6} & 89.2 & 89.4 \\
    WinoGrande & 5-shot & 83.8 & 85.3 & \textbf{85.9} & 85.6 \\
    TriviaQA & 5-shot & 80.3 & \textbf{85.1} & 83.5 & 83.9 \\
    GPQA-Diamond & 5-shot & \textbf{55.1} & 48.1 & 51.0 & 52.0 \\
    SuperGPQA & 5-shot & 41.1 & \textbf{44.7} & 42.3 & 43.6 \\
    SimpleQA & 5-shot & 20.6 & \textbf{35.3} & 26.3 & 27.0 \\
    \midrule
    \multicolumn{6}{l}{\textbf{Mathematics}} \\
    GSM8K & 8-shot & \textbf{92.3} & 92.1 & 91.4 & 91.1 \\
    MATH & 4-shot & \textbf{71.0} & 70.2 & 62.6 & 62.5 \\
    AIME 24\&25 & 2-shot & \textbf{35.3} & 31.6 & 21.6 & 24.8 \\
    \midrule
    \multicolumn{6}{l}{\textbf{Code}} \\
    HumanEval+ & 1-shot & 70.7 & \textbf{84.8} & 64.6 & 67.7 \\
    MBPP+ & 3-shot & 71.4 & \textbf{73.8} & 72.2 & 69.8 \\
    CRUXEval-I & 1-shot & 67.5 & \textbf{74.0} & 62.1 & 63.9 \\
    CRUXEval-O & 1-shot & 79.1 & \textbf{83.5} & 76.4 & 74.9 \\
    MultiPL-E HumanEval & 0-shot & 59.5 & \textbf{60.5} & 45.9 & 45.7 \\
    MultiPL-E MBPP & 0-shot & 56.7 & \textbf{58.8} & 52.5 & 50.6 \\
    BigCodeBench & 0-shot & \textbf{70.1} & 61.7 & 63.0 & 62.9 \\
    LiveCodeBench v6 & 1-shot & \textbf{30.8} & 26.3 & 24.8 & 24.9 \\
    SWE-Bench {\tiny (AgentLess Repair)} & 3-shot & \textbf{30.8} & 28.2 & 24.8 & 9.4$^*$ \\
    \midrule
    \multicolumn{6}{l}{\textbf{Chinese}} \\
    C-Eval & 5-shot & 87.9 & \textbf{92.5} & 90.0 & 91.0 \\
    CMMLU & 5-shot & 87.4 & \textbf{90.9} & 88.8 & 88.9 \\
    C-SimpleQA & 5-shot & 61.5 & \textbf{77.6} & 70.9 & 68.0 \\
    \midrule
    \multicolumn{6}{l}{\textbf{Multilingual}} \\
    GlobalMMLU & 5-shot & 76.6 & 80.7 & 81.9 & \textbf{82.0} \\
    INCLUDE & 5-shot & 71.4 & 75.3 & 77.2 & \textbf{77.2} \\
    \bottomrule
    \end{tabular}
    }
    \caption{
        Comparison among \mimoflash{} and other open-source base models. An asterisk (*) denotes that the model does not follow the format of few-shot examples.
    }
    \label{tab:base_eval}
\end{table}

Table~\ref{tab:base_eval} presents a comprehensive comparison of \mimoflash{}-Base against leading open-source base models~\citep{team2025kimik2,liu2024deepseek}.
\mimoflash{}-Base delivers competitive performance across most benchmarks and consistently outperforms peers on reasoning tasks (MMLU-Pro, GPQA-Diamond, AIME).
On SWE-Bench, it even surpasses the substantially larger Kimi-K2-Base while using less than one-third the parameters, underscoring the strength of our approach for realistic code-agent tasks. 
However, constrained by its limited parameter count, we observe \mimoflash{} exhibits lower knowledge capacity compared to larger models, as reflected in SimpleQA.

\begin{table}[htbp]
    \centering
    \small
    \resizebox{0.99\textwidth}{!}{
    \begin{tabular}{l c | c | c c c}
    \toprule
    \multirow{2}{*}{\textbf{Benchmark}} & \multirow{2}{*}{\textbf{Length}} & \textbf{MiMo-V2-Flash} & \textbf{Kimi-K2} & \textbf{DeepSeek-V3.1} & \textbf{\textbf{DeepSeek-V3.2}} \\
    & & \textbf{Base} & \textbf{Base} & \textbf{Base} & \textbf{Exp Base} \\
    \midrule
    \#Activated Params & - & 15B & 32B & 37B & 37B \\
    \#Total Params & - & 309B & 1043B & 671B & 671B \\
    \midrule
    \multirow{4}{*}{NIAH-Multi} & 32K & 99.3 & 99.8 & 99.7 & 85.6$^*$ \\
     & 64K & 99.9 & 100.0 & 98.6 & 85.9$^*$ \\
     & 128K & 98.6 & 99.5 & 97.2 & 94.3$^*$ \\
     & 256K & 96.7 & - & - & - \\
    \midrule
    \multirow{4}{*}{GSM-Infinite Hard} & 16K & 37.7 & 34.6 & 41.5 & 50.4 \\
     & 32K & 33.7 & 26.1 & 38.8 & 45.2 \\
     & 64K & 31.5 & 16.0 & 34.7 & 32.6 \\
     & 128K & 29.0 & 8.8 & 28.7 & 25.7 \\
    \bottomrule
    \end{tabular}
    }
    \caption{
        Long context performance of \mimoflash{} and other open-source base models. An asterisk (*) indicates the model may fail to follow the prompt. All baseline models have maximum model lengths shorter than 256K.
    }
    \label{tab:long_eval}
\end{table}

We illustrate the long context capabilities of each model in Table~\ref{tab:long_eval}.
For long-context retrieval, our model architecture achieves a near 100\% success rate from 32K to 256K.
On the extreme stress long context reasoning benchmark GSM-Infinite, \mimoflash{} also shows strong performance, with minimal performance degradation from 16K to 128K.
In contrast, DeepSeek-V3.2-Exp, a sparse attention LLM, attains the highest score under 32K but degrades substantially at 64K and 128K, suggesting an intrinsic disadvantage in long-context reasoning with noisy inputs.
These results strongly demonstrate the effectiveness and scalability of our hybrid SWA architecture, vanilla 32K pretraining, and context extension training.

\section{Post-Training}

\subsection{Multi-Teacher On-Policy Distillation (MOPD): A New Post-Training Paradigm}
\label{subsec:mopd-overview}

\begin{figure*}[t]
    \centering
    \includegraphics[width=\linewidth]{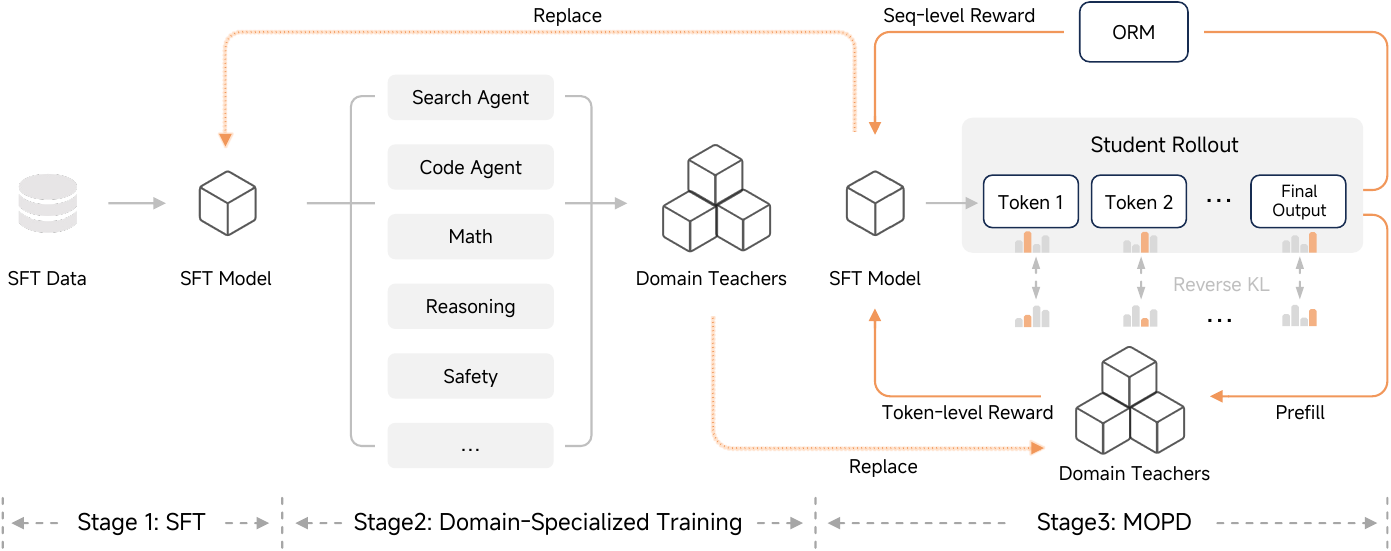}
    \caption{Overview of MiMo-V2-Flash post-training stages.}
    \label{fig:post-training}
\end{figure*}

Modern language models increasingly rely on extensive post-training to enhance their intelligence and capabilities. However, current post-training pipelines face fundamental challenges: \textit{capability imbalance}, where improving one skill causes regressions in others (the ``see-saw'' effect), and \textit{learning inefficiency}, where existing approaches fail to fully leverage training signals when combining knowledge from multiple specialized models.

We propose Multi-Teacher On-Policy Distillation (MOPD), a unified post-training paradigm that addresses these challenges through a three-stage framework, as illustrated in Figure~\ref{fig:post-training}:

\paragraph{Stage 1: Supervised Fine-Tuning (SFT)} We establish foundational instruction-following capabilities through supervised learning on high-quality instruction-response pairs, enabling the model to understand and execute user requests across diverse domains.

\paragraph{Stage 2: Domain-Specialized Training} We train a suite of domain-specialized teacher models through independent RL optimization on focused tasks including agentic capabilities (search, coding, general tool use) and non-agentic tasks (mathematical reasoning, general reasoning, safety alignment). Each teacher achieves superior performance in its respective domain through targeted optimization with domain-specific reward signals.

\paragraph{Stage 3: Multi-Teacher On-Policy Distillation} Rather than merging model parameters or generating static offline datasets from experts, we formulate multi-teacher knowledge integration as an on-policy reinforcement learning process. The student model samples from its own evolving distribution and receives token-level supervision from domain-specific teachers through KL divergence rewards~\citep{minillm,Agarwal2023OnPolicyDO,lu2025onpolicydistillation}, effectively combining specialized capabilities without the traditional trade-offs~(Table~\ref{tab:mopd_eval}).

\begin{table}[htbp]
    \centering
    \small
    \begin{tabular}{l c c c c}
    \toprule
    \multirow{2}{*}{\textbf{Benchmark}}  & \textbf{Student} & \multirow{2}{*}{\textbf{Best Teacher}} & \textbf{Student} & \multirow{2}{*}{\textbf{$\Delta$(Student-Teacher)}} \\
                      & \textbf{Before MOPD} &  & \textbf{After MOPD} &  \\
    \midrule
AIME 2025                          &  89.3  &  93.9 \tiny{(RL)}  &  94.1 & $+$0.2 \\
HMMT Feb. 2025                          &  76.9  &  82.6 \tiny{(RL)}  &  84.4 & $+$1.8 \\
LiveCodeBench                &  77.5  &  82.6 \tiny{(RL)}  &  83.2 & $+$0.6 \\
MMLU-Pro                        &  84.7  &  84.7 \tiny{(Self)}  &  84.9 & $+$0.2 \\
GPQA-Diamond                    &  84.9  &  84.9 \tiny{(Self)}  &  84.3 & $-$0.6 \\
HLE \tiny{(w/o Tool)}                  &  21.2  &  21.2 \tiny{(Self)}  &  22.1 & $+$0.9 \\
Arena-Hard \tiny{(Hard Prompt)}        &  50.0  &  50.0 \tiny{(Self)}  &  54.1 & $+$4.1 \\
Arena-Hard \tiny{(Creative Writing)}   &  90.1  &  90.1 \tiny{(Self)}  &  86.2 & $-$3.9 \\
SWE-Bench Verified              &  67.8  &  74.2 \tiny{(RL)}  &  73.4 & $-$0.8 \\
Tau2-Bench                      &  75.9  &  79.6 \tiny{(RL)}  &  80.3 & $+$0.7 \\
Tau2-Bench \tiny{(Telecom)}              &  92.7  &  95.0 \tiny{(RL)}  &  95.3 & $+$0.3 \\
BrowseComp                      &  42.5  &  51.7 \tiny{(SFT)}  &  45.4 & $-$6.3 \\
    \bottomrule
    \end{tabular}
    \caption{Benchmark results of MOPD. The model types of best teachers are tagged, including RL, SFT, and the student model itself.}
    \label{tab:mopd_eval}
\end{table}

This unified framework offers several critical advantages over traditional post-training approaches:

\begin{itemize}
    \item \textit{Effective and Efficient.} 
    Unlike parameter merging or sequential training, which often trade off capabilities, MOPD preserves peak performance of the strongest teacher across all domains. Furthermore, on-policy distillation using dense, token-level rewards from teacher logits ensures stable credit assignment and rapid convergence. By learning from its own distribution, the student avoids the exposure bias and distribution mismatch common in off-policy methods trained on static datasets.
    \item \textit{Modular and Scalable.} The choice of teacher model is highly flexible: it can be a specialized RL-derived model with strong capabilities, a different SFT model, or even the student model itself. The decoupled design enables easy integration of new teachers without restructuring the entire pipeline. Moreover, the framework works seamlessly with existing outcome reward models (ORMs) and is especially advantageous for complex agentic tasks, where setting up independent training pipelines would otherwise be cumbersome.
    \item \textit{Iterative Co-Evolution.} MOPD naturally supports a teacher-student co-evolution cycle. Distilled student models can re-enter the specialized RL stage to produce stronger teachers, which in turn provide higher-quality supervision for the next generation of students, forming a self-reinforcing improvement cycle that enables sustained capability scaling.
\end{itemize}

In the following subsections, we detail each stage of the MOPD paradigm, beginning with supervised fine-tuning (\S\ref{subsec:post_train_sft}), followed by specialized RL for both agentic and non-agentic tasks~(\S\ref{subsec:scale_rl}), and conclude with the technical formulation of the multi-teacher distillation mechanism (\S\ref{subsec:mopd-technical}).

\subsection{Supervised Fine-Tuning (SFT)}
\label{subsec:post_train_sft}

The SFT stage serves as the foundation of our post-training pipeline, transforming the base model into a helpful assistant capable of following instructions and responding effectively across diverse tasks. This stage is crucial for activating the model's latent capabilities acquired during pre-training and aligning its outputs with desired formats and styles.

To achieve this, we curated millions of training samples spanning diverse domains, including general conversation, reasoning, coding, and agent tasks. These samples cover both thinking and non-thinking modes, with responses generated by our in-house domain-specialized model checkpoints. This diverse training mixture ensures comprehensive capability activation across the model's intended use cases.

Through preliminary experiments, we identified a critical stability metric for MoE SFT training: the number of parameters with zero gradients (num-zeros). This metric provides early warning signals for training instability: an increasing num-zeros indicates deteriorating load balance among experts, while a decreasing num-zeros suggests the model is significantly overfitting to the training data. Maintaining stable num-zeros throughout training is therefore essential for successful SFT. Furthermore, this stability is paramount for ensuring the robustness and convergence of the subsequent RL phase.

Our experiments reveal that num-zeros stability critically depends on two hyperparameters: the expert bias update rate and the $\epsilon$ parameter in the AdamW optimizer. Based on these findings, we configure our training with the following hyperparameters. We employ a cosine decay learning rate scheduler from $5.0 \times 10^{-5}$ to $5.0 \times 10^{-6}$, with a batch size of 128 and AdamW $\epsilon$ set to $1.0 \times 10^{-8}$. The MoE expert bias update rate is set to $1.0 \times 10^{-4}$, and the sequence auxiliary loss coefficient to $1.0 \times 10^{-6}$.

\subsection{Scaling Reinforcement Learning (RL)}
\label{subsec:scale_rl}

Reinforcement learning pushes model capabilities beyond what supervised fine-tuning alone can achieve. 
We employ different RL strategies depending on whether tasks involve agentic behavior, scaling both non-agentic and agentic RL training to maximize performance across diverse domains.

\subsubsection{Non-Agentic RL Training}

Non-agentic RL training focuses on improving the model's performance on single-turn tasks, where the model generates a complete response without requiring interactive feedback or multi-step execution. The primary objective is to enhance the model's reasoning accuracy in verifiable domains (e.g., mathematics, coding, logic) while simultaneously aligning its outputs for helpfulness and safety in open-ended conversations.

Our approach to generating reward signals varies based on task characteristics. For domains with verifiable outcomes, we employ a hybrid verification system combining programmatic tools with an LLM judge to automatically assess correctness against curated problem-solution pairs. 
For subjective qualities such as helpfulness and safety, we implement a rubric-based framework where an advanced LLM judge evaluates responses against detailed rubrics and reference answers, producing granular reward signals that guide the model toward desired behaviors.

\subsubsection{Agentic RL Training}

\begin{table}[htbp]
\small
    \centering
    \begin{tabular}{lccc}
        \toprule
        \textbf{Agent Type} & \textbf{Number of Tasks} & \textbf{Environment} & \textbf{Prompt Source} \\
        \midrule
        Code Agent & $90\textrm{K}$ & Real & Real  \\
        Code Agent & $30\textrm{K}$ & Real & Synthesized \\
        Search Agent & $150\textrm{K}$ & Real & Synthesized \\
        General Agent & $50\textrm{K}$ & Synthesized & Synthesized \\
        \bottomrule
    \end{tabular}
    \caption{A summary of our training data composition across different agent types. We leverage both real-world and synthetically generated data to create a diverse set of tasks for training agents in various environments.}
    \label{tab:agent_tasks}
    
\end{table}
While non-agentic RL focuses on single-turn reasoning and generation, agentic RL trains the model to operate in interactive, multi-turn environments requiring planning, action execution, and adaptation based on feedback. We scale agentic RL along two critical dimensions: environment diversity and compute resources.

\paragraph{Scaling Agentic Environment Diversity} We construct a diverse suite of agentic training environments spanning code debugging, terminal operations, web development, and general tool use (Table~\ref{tab:agent_tasks}). Each environment targets distinct capabilities while sharing the common requirement of multi-step reasoning and execution. Below, we elaborate on the details for agentic environments.

\paragraph{Code Agent} We train on large-scale code agentic tasks derived from real-world GitHub issues, where the model operates in an agentic loop to read and edit files, execute commands, and receive rewards based on verifiable unit tests. Our key insight is that continuously scaling available tasks drives sustained improvements in code intelligence. To enable efficient RL training on over 100,000 code tasks, we develop two infrastructure components. First, we build an automated environment setup pipeline that provisions development environments from repository snapshots and packages them into containerized images, achieving 70\% success rate across 8 programming languages and supported by a large-scale Kubernetes cluster running over 10,000 concurrent pods. Second, we implement a lightweight agent scaffold that integrates seamlessly with Kubernetes, Docker, or local backends, exposing three atomic tools (\texttt{bash}, \texttt{str\_replace}, \texttt{finish}) that interact with execution backends solely via shell commands. This design eliminates server-based tool implementations and employs a minimal system prompt without predefined workflows, allowing the model to discover best practices during training.

\paragraph{Terminal Agent} Beyond GitHub issues, we strengthen terminal-based problem-solving capabilities using tasks sourced from Stack Overflow and Stack Exchange. We select materials requiring advanced technical expertise and transform them into computational tasks with corresponding queries, Dockerfiles, and test cases. After verifying environment installation and filtering by difficulty and reliability, we obtain approximately 30,000 queries with validated execution environments. Additional filtering based on pass rates removes tasks with unreliable correctness judgments or insufficient complexity for effective RL training.

\paragraph{Web Development Agent} To improve web development code generation, we build a real-world-grounded synthetic dataset paired with a multimodal verifier. We collect high-quality user-written web pages, execute generated code using Playwright to obtain rendered videos, and apply a multimodal visual discriminator to retain only high-quality samples, where video-based evaluation reduces visual hallucination compared to static screenshots. We reverse-engineer user queries from curated pages as seed prompts to synthesize large-scale RL data covering eight web categories that closely match real-world usage. Our vision-based verifier scores rollout executions from recorded videos, jointly evaluating visual quality, functional correctness, and executability to ensure rewards reflect both appearance and behavior.

\paragraph{General Agent} We develop two general agentic capabilities. Our search agent adopts a scaffold providing three core tools (search, open, find) for autonomous web exploration. We construct queries through recursive fact-graph expansion from seed entities, where difficulty scales with relation chain depth and detail obfuscation, enabling automated generation of challenging search problems with verifiable answers. Our function-calling agent trains on synthetic application environments with custom toolsets constructed by generating tool-call graphs based on explicit data dependencies (direct input-output relationships) and implicit logical dependencies (reasoning about hidden system states), requiring both data propagation and state inference capabilities.

\begin{figure*}[t!]
    \centering
    \includegraphics[width=0.95\linewidth]{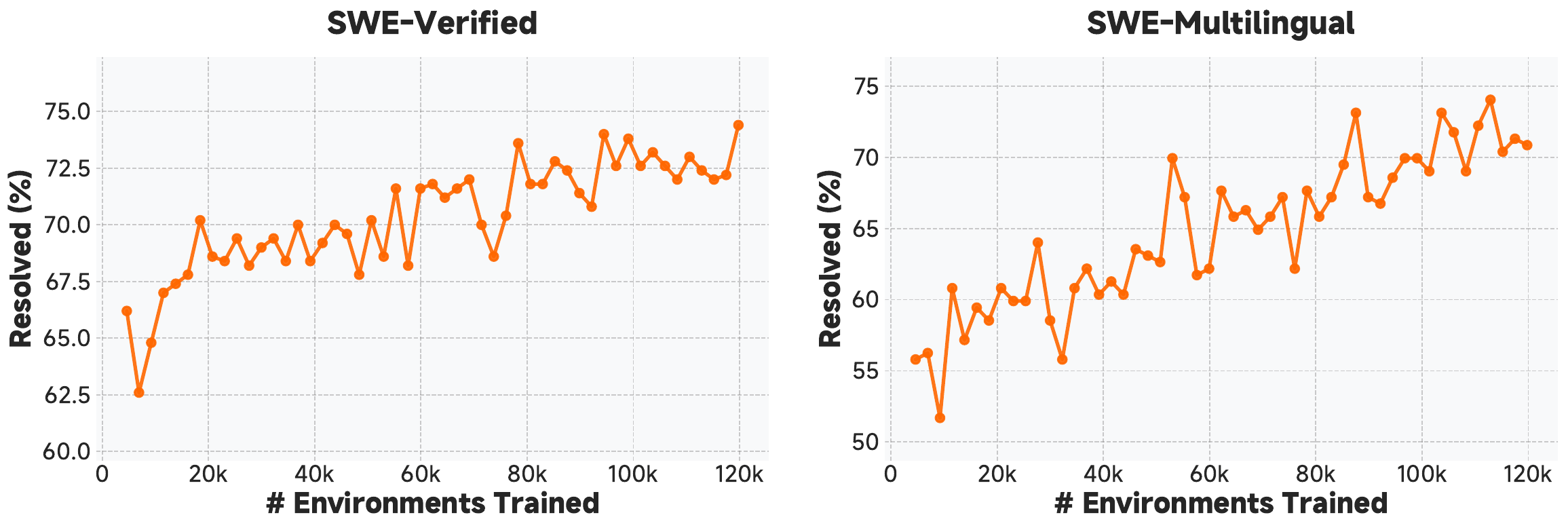}
    \caption{Code-agentic RL scaling curves. The X-axis represents total interactive environments consumed during on-policy RL rollouts; the Y-axis shows resolved rates on SWE-Bench-Verified and SWE-Bench-Multilingual.}
    \label{fig:swebench_scaling}
\end{figure*}

\begin{figure*}[t!]
    \centering
    \includegraphics[width=1.0\linewidth]{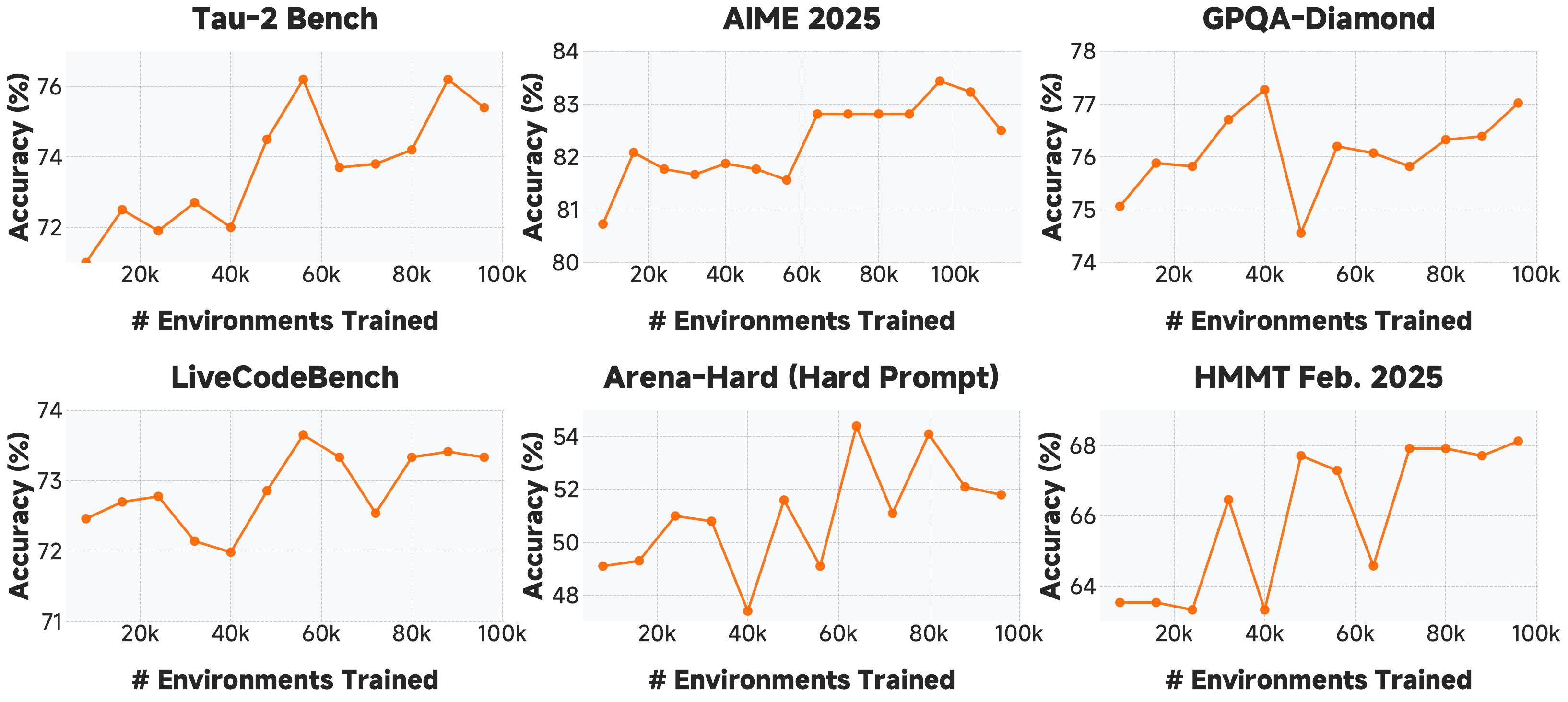}
    \caption{Generalization of code-agentic RL training to other task domains.}
    \label{fig:swe_generalize}
\end{figure*}

\paragraph{Scaling Agentic Compute}
Training on the previous diverse set of agentic environments (Table~\ref{tab:agent_tasks}), we find that scaling agentic RL compute not only boosts code-agentic performance but also generalizes effectively to other task types. Figure~\ref{fig:swebench_scaling} shows the RL training curve for our code-agent, where the model performed on-policy rollouts and updates across approximately 120K environments. 
This scaling significantly improves upon the SFT base model's performance on SWE-Bench-Verified and SWE-Bench-Multilingual. Moreover, Figure~\ref{fig:swe_generalize} demonstrates that large-scale code-agentic RL training generalizes effectively to other agentic tasks, as well as math, code, and general reasoning benchmarks, suggesting that agentic training develops broadly transferable problem-solving capabilities.

\subsection{Technical Formulation of MOPD}
\label{subsec:mopd-technical}

Having established the foundation through SFT and trained specialized teachers through domain-specific RL, we now formalize the multi-teacher on-policy distillation mechanism that integrates these specialized capabilities into a unified student model.

Specifically, we cast multi-teacher distillation as an on-policy reinforcement learning objective. Let $\pi_\theta$ denote the target student policy optimized in the training engine, $\mu_\theta$ denote the student sampling policy adopted in the inference engine, and $\pi_{\text{domain}_x}$ denote the teacher policy specialized for the domain of prompt $x$ sampled from distribution $\mathcal{D}$. Let $\text{sg}[\cdot]$ denote the stop-gradient operator. The reverse KL divergence loss between student and teacher is defined as:
\begin{equation}
    \mathcal{L}_\text{reverse-KL}(\theta)
  =-\mathbb{E}_{x\sim\mathcal{D},  y_{t}\sim \pi_\theta(\cdot|x, y_{<t})}\log\frac{ \pi_{\text{domain}_x}(y_{t}|x, y_{<t})}{ \pi_\theta(y_{t}|x, y_{<t})}.
\end{equation}
The gradient is:
\begin{equation}
    \nabla_{\theta}\mathcal{L}_\text{reverse-KL}(\theta)
  =-\mathbb{E}_{x\sim\mathcal{D},  y_{t}\sim \pi_\theta(\cdot|x, y_{<t})}\left[\log\frac{ \pi_{\text{domain}_x}(y_{t}|x, y_{<t})}{ \pi_\theta(y_{t}|x, y_{<t})} \nabla_\theta \log \pi_\theta(y_{t}|x, y_{<t})\right].
\end{equation}
Following~\citet{IcePop2025}, we apply training-inference importance sampling and discard tokens that exhibit large discrepancies. We then define the surrogate loss of MOPD as:
\begin{equation}
\mathcal{L}_{\text{MOPD}}(\theta) = -\mathbb{E}_{x \sim\mathcal{D}, y \sim \mu_{\theta}(\cdot|x)}
\left[\frac{1}{|y|}\sum_{t=1}^{|y|}
  w_{t} \hat{A}_{\text{MOPD},t}\log\pi_{\theta}(y_{t}|x, y_{<t}) \right],
\end{equation}
where
\begin{equation}
    w_{t}(\theta)=
\begin{cases}
\text{sg}\left[\frac{\pi_\theta(y_{t}|x, y_{<t})}{\mu_\theta(y_{t}|x, y_{<t})}\right], & \epsilon_{\text{low}} \le \frac{\pi_\theta(y_{t}|x, y_{<t})}{\mu_\theta(y_{t}|x, y_{<t})} \le \epsilon_{\text{high}}, \\
0,  & \text{other},
\end{cases}
\quad
    \ \hat{A}_{\text{MOPD},t} = \text{sg}\left[\log\frac{ \pi_{\text{domain}_x}(y_{t}|x, y_{<t})}{ \pi_\theta(y_{t}|x, y_{<t})}\right].
\label{eq:advantage_calculation}
\end{equation}

By default, we combine the advantages of MOPD with other types of advantages, such as those computed using Outcome Reward Models (ORMs), including GRPO~\citep{shao2024deepseekmath}. 
Let $\hat{A}_{\text{ORM}}$ denote the advantages computed by the ORMs; the final advantages are given by:
\begin{equation}
    \hat{A}_{\text{MOPD},t} = \text{sg}\left[\log\frac{ \pi_{\text{domain}_x}(y_{t}|x, y_{<t})}{ \pi_\theta(y_{t}|x, y_{<t})}\right] + \alpha \hat{A}_{\text{ORM}}.
\end{equation}

\begin{figure}[t]
    \centering
    \includegraphics[width=0.48\linewidth]{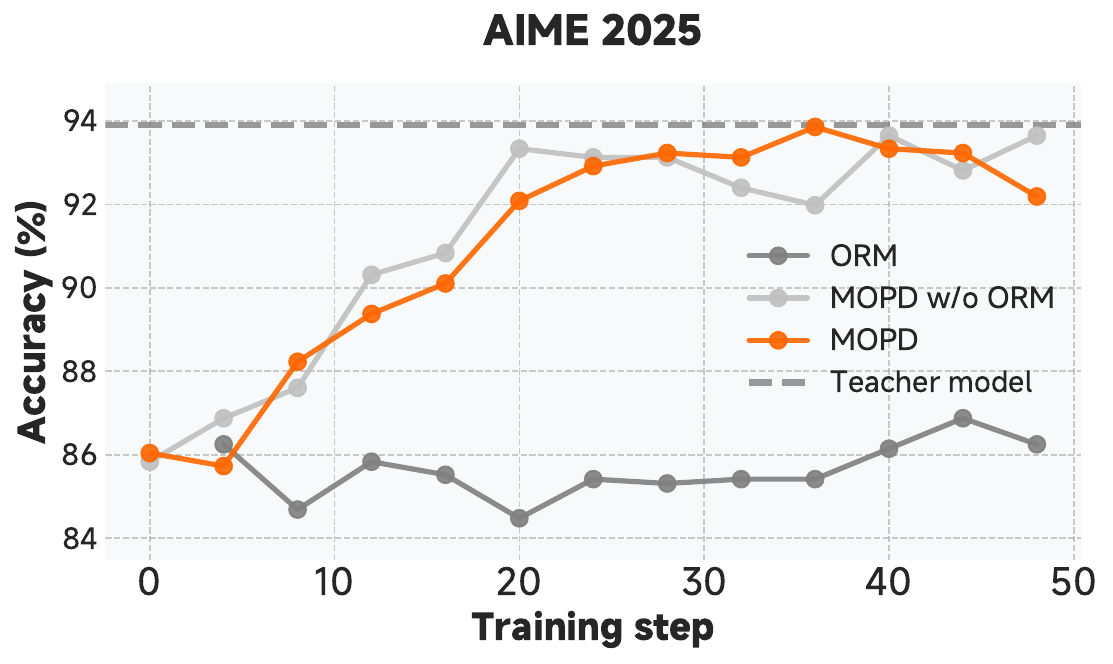}
    \includegraphics[width=0.48\linewidth]{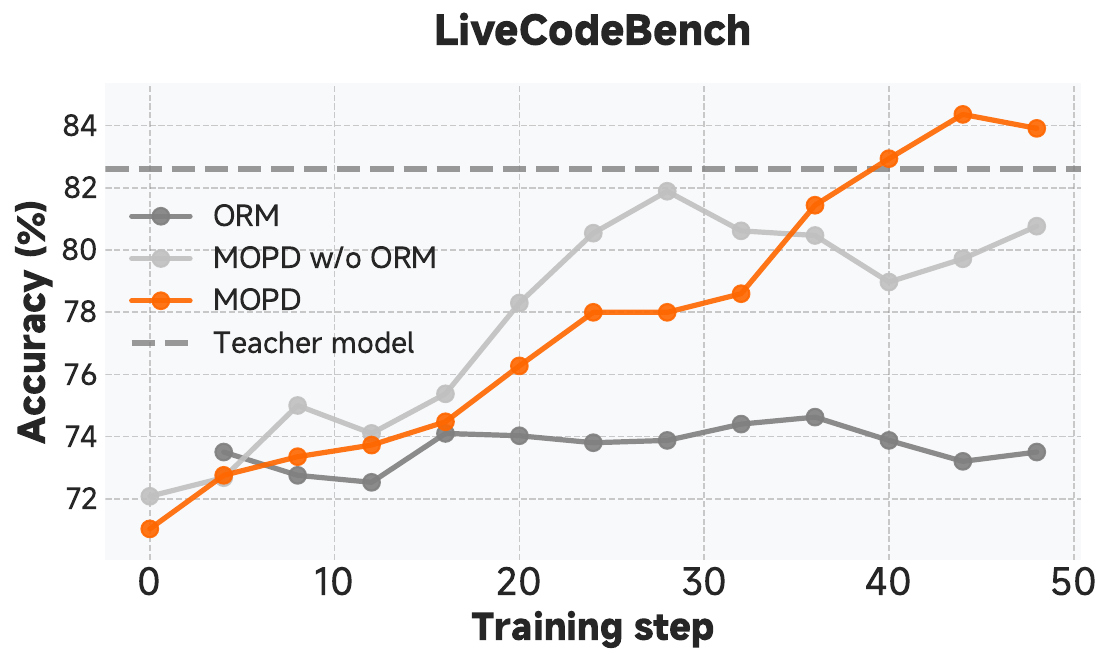}
    \caption{Comparison of different post-training methods on math and code tasks. Three lines represent training RL with ORM, MOPD without outcome rewards (MOPD w/o ORM), and MOPD. }
    \label{fig:compare-post-train}
\end{figure}

Figure~\ref{fig:compare-post-train} demonstrates the effectiveness of MOPD compared to traditional post-training approaches. On mathematical reasoning (AIME 2025) and coding (LiveCodeBench) benchmarks, MOPD successfully preserves and combines specialized capabilities from multiple teachers, achieving performance that matches or exceeds the strongest teacher in most domains.

\subsection{Evaluations}
\label{subsec:post_train_eval}

\subsubsection{Evaluation Setup}

We evaluate \mimoflash{} on MMLU-Pro~\citep{wang2024mmlu}, GPQA-Diamond~\citep{rein2024gpqa}, HLE Text-only~\citep{phan2025humanity}, AIME 2025~\citep{AIME}, LiveCodeBench (2024.08-2025.04)~\citep{jain2024livecodebench},  HMMT Feb. 2025~\citep{balunovic2025matharena}, Arena-Hard~\citep{arenahard2024}, LongBench V2~\citep{bai2025longbench}, MRCR~\citep{vodrahalli2024michelangelo} (\{2,4,8\}-needles, maximum 128K), SWE-Bench Verified~\citep{jimenez2024swebench}, SWE-Bench Multilingual~\citep{yang2025swesmith}, Terminal-Bench, BrowseComp~\citep{wei2025browsecomp}, $\tau^2$-Bench~\citep{barres2025tau}.

\subsubsection{Evaluation Results}

We illustrate the evalution results in Table~\ref{tab:main_eval}.
\mimoflash{} achieves performance comparable to that of Kimi-K2-Thinking and DeepSeek-V3.2-Thinking on most reasoning benchmarks.
The model also maintains competitive general writing capabilities, enabling it to generate high-quality responses on open-ended tasks.
In long context evaluations, our model surpasses Kimi-K2-Thinking, a significantly larger full global attention LLM, highlighting the strong long-context capabilities of our hybrid SWA architecture.

Notably, \mimoflash{} achieves 73.4\% on SWE-Bench Verified, outperforming all open-source competitors and approaching the performance of GPT-5-High.
On SWE-Bench Multilingual, our model resolves 71.7\% issues, establishing it as the most capable open-source LLM for software engineering tasks.
These results underscore the effectiveness of our ultra-scaled agentic RL training. On Terminal Bench, the model also delivers a competitive score. 

In search agent evaluation, \mimoflash{} scores 45.4 on BrowseComp, and is further boosted to 58.3 with the context management method outlined in Appendix~\ref{sec:context_management}.
For general tool-use on $\tau^2$-Bench, we employ DeepSeek-V3.2 as the user agent, achieving category scores of 95.3 (Telecom), 79.5 (Retail), 66.0 (Airline).

Taken together, these results validate the effectiveness of our ultra-large-scale RL training within the MOPD post-training paradigm, and highlight the models strong potential for real-world coding, reasoning, and agentic workflows.

\begin{table}[htbp]
    \centering
    \small
    \resizebox{0.99\textwidth}{!}{
    \begin{tabular}{l | c | c c c c c}
    \toprule
    \multirow{2}{*}{\textbf{Benchmark}} & \textbf{MiMo-V2} & \textbf{Kimi-K2} & \textbf{\textbf{DeepSeek-V3.2}} & \textbf{Gemini-3.0} & \textbf{Claude} & \textbf{GPT-5} \\
    & \textbf{Flash} & \textbf{Thinking} & \textbf{Thinking} & \textbf{Pro} & \textbf{Sonnet 4.5} & \textbf{High} \\
    \midrule
    \multicolumn{6}{l}{\textbf{Reasoning}} \\
    MMLU-Pro & 84.9 & 84.6 & 85.0 & 90.1 & 88.2 & 87.5 \\
    GPQA-Diamond & 84.3 & 84.5 & 82.4 & 91.9 & 83.4 & 85.7 \\
    HLE {\tiny (no tools)}  & 22.1 & 23.9 & 25.1 & 37.5 & 13.7 & 26.3 \\
    AIME 2025 & 94.1 & 94.5 & 93.1 & 95.0 & 87.0 & 94.6 \\
    HMMT Feb. 2025 & 84.4 & 89.4 & 92.5 & 97.5 & 79.2 & 88.3 \\
    LiveCodeBench-v6 & 85.1 & 83.1 & 83.3 & 90.7 & 64.0 & 84.5 \\
    \midrule
    \multicolumn{6}{l}{\textbf{General Writing}} \\
    Arena-Hard {\tiny (Hard Prompt)} & 54.1 & 71.9 & 53.4 & 72.6 & 63.3 & 71.9 \\
    Arena-Hard {\tiny (Creative Writing)} & 86.2 & 80.1 & 88.8 & 93.6 & 76.7 & 92.2 \\
    \midrule
    \multicolumn{6}{l}{\textbf{Long Context}} \\
    LongBench V2 & 60.6 & 48.1 & 58.4 & 65.6 & 61.8 & - \\
    MRCR & 45.7 & 44.2 & 55.5 & 89.7 & 55.4 & - \\
    \midrule
    \multicolumn{6}{l}{\textbf{Code Agent}} \\
    SWE-Bench Verified & 73.4 & 71.3 & 73.1 & 76.2 & 77.2 & 74.9 \\
    SWE-Bench Multilingual & 71.7 & 61.1 & 70.2 & - & 68.0 & 55.3 \\
    Terminal-Bench Hard &30.5  &30.6 & 35.4 & 39.0 & 33.3 & 30.5 \\
    Terminal Bench 2.0 &38.5  & 35.7 & 46.4 & 54.2 & 42.8 & 35.2 \\
    \midrule
    \multicolumn{6}{l}{\textbf{General Agent}} \\
    BrowseComp & 45.4 & - & 51.4 & - & 24.1 & 54.9 \\
    BrowseComp {\tiny (w/ Context Manage)} & 58.3 & 60.2 & 67.6 & 59.2 & - & - \\
    $\tau^2$-Bench & 80.3 & 74.3 & 80.3 & 85.4 & 84.7 & 80.2  \\
    \bottomrule
    \end{tabular}
        }
    \caption{
        Comparison between \mimoflash{} and open/closed models.
    }
    \label{tab:main_eval}
\end{table}

\subsection{RL Infrastructures}

Our RL (and MOPD) infrastructure uses SGLang~\citep{zheng2024sglang} as the inference engine and Megatron-LM~\citep{shoeybi2019megatron} as the training engine.
We adopt FP8 for both training and inference.
To enable stable, efficient, and flexible RL training, we implement three extended modules: Rollout Routing Replay~\citep{ma2025stabilizing} (Sec~\ref{sec:posttrain:rlinfra:r3}), Data Scheduler (Sec~\ref{sec:posttrain:rlinfra:data}), and Tool Manager combined with Toolbox (Sec~\ref{sec:posttrain:rlinfra:tool}).

\subsubsection{Stablized Training via Rollout Routing Replay (R3)}
\label{sec:posttrain:rlinfra:r3}

MoE models suffer from inconsistent expert routing across rollout and training due to numerical precision issues~\citep{he2025nondeterminism,yao2025offpolicy}.
We propose Rollout Routing Replay (R3)~\citep{ma2025stabilizing} to train RL using the same routed experts from rollout, making its overhead negligible through optimized data types and communication overlapping.
For multi-turn agent training, we employ a request-level prefix cache during rollout.
This cache stores KVCache and MoE routed experts from prior turns, allowing them to be reused for subsequent generation steps of the same request.
Unlike the commonly-used radix cache in current inference engines, our request-level prefix cache avoids re-prefilling or inter-request output cache sharing, ensuring sampling consistency for routed experts.

\subsubsection{Data Scheduler}
\label{sec:posttrain:rlinfra:data}

For MiMo-V2-Flash, we extend the Seamless Rollout Engine~\citep{xia2025mimo} and implement a Data Scheduler to seamlessly schedule fine-grained sequences instead of micro-batches, addressing GPU idleness in distributed MoE training.
In dynamic sampling, as sequences return for reward computation, we reference historical pass rates and, if necessary, assign new prompts to GPUs with load balancing.
We integrate partial rollout~\citep{team2025kimi,fu2025areal} to partition overlong trajectories across steps, while limiting staleness and the proportion of partial samples in each batch.
By employing staleness-aware truncated importance sampling for partial rollout, we significantly accelerate RL training without sacrificing model quality.

The Data Scheduler supports data source-specific configurations (sample quota, scheduling priority, length limits, temperature) and fits pass rates to accept samples by configured ratios.
Priority-based scheduling overlaps reward computation and inference across data sources with different time patterns, ensuring high GPU utilization.

\subsubsection{Toolbox and Tool Manager}
\label{sec:posttrain:rlinfra:tool}

We implement Toolbox and Tool Manager to tackle global resource contention and local inefficiency in RL agent training.
These modules leverage Ray~\citep{moritz2018ray} for efficient scheduling.
Toolbox acts as the centralized resource allocator, enforcing resource quota and QPS limits for tools across concurrent tasks.
It adopts fault-tolerant Ray actor pools, which eliminate cold-start delays.
Integrated with the rollout engine, Tool Manager coordinates with Toolbox to accelerate training through environment pre-warming and sequence-level asynchronous reward computation.
It maintains training stability through timeout recovery and real-time monitoring.
By disaggregating the tool management and rollout workflow, Toolbox isolates task-specific logic from system-wide policies, enabling modular extensibility without compromising stability.

\section{MTP Speedup}
\label{sec:mtp}

\subsection{MTP Acceptance Length}
\begin{figure*}[t!]
    \centering
    \includegraphics[width=0.6\linewidth]{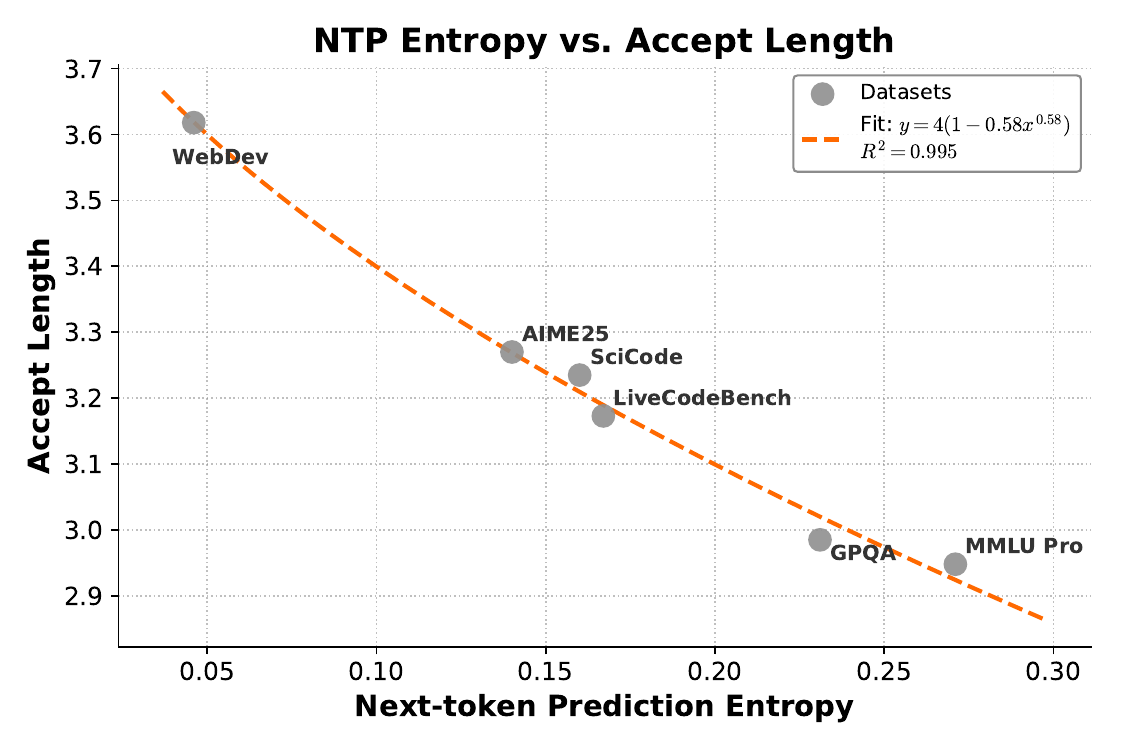}
    \caption{The correlation between next token cross-entropy and Average Accept Length across different datasets. The orange dashed line represents the best-fit curve ($R^2=0.995$).}
    \label{fig:entropy_vs_accept_length}
\end{figure*}

We analyze the relationship between the model's predictive uncertainty measured by next
token cross-entropy and the efficiency of the Multi-Token Prediction (MTP) module. As illustrated in Figure~\ref{fig:entropy_vs_accept_length}, we evaluate the average acceptance length with 3 MTP layers across diverse benchmarks, ranging from code generation (e.g., WebDev, LiveCodeBench) to complex reasoning tasks (e.g., AIME25, MMLU Pro). 

The results reveal a strong inverse correlation: lower entropy contexts (such as WebDev) allow for significantly longer acceptance sequences, reaching approximately 3.6 tokens. Conversely, tasks with higher intrinsic uncertainty (e.g., MMLU Pro) exhibit shorter acceptance lengths due to increased prediction divergence. This behavior is accurately modeled by a log-transformed fit ($y = 4(1 - 0.58x^{0.58})$) with an $R^2$ of $0.995$, suggesting that next token cross-entropy is a primary determinant of MTP throughput.

\subsection{MTP Inference Speedup}

We measure the decoding speedup of \mimoflash{} with 3-layer MTP across varying batch sizes (per node) and accept lengths, using 16K input and 1K output lengths. 
The results in Table~\ref{tab:mtp_speedup} demonstrate that MTP consistently outperforms the baseline without additional hardware costs.
Notably, the speedup scales linearly with accept length.
Under different batch sizes, MTP exhibits varying speedup, which depends on the corresponding computation and I/O demands as well as kernel efficiency.
In practice, researchers and engineers should tune both batch size and MTP layers based on hardware roofline models to optimize the speed-cost trade-off.

\begin{table}[t!]
\small
    \centering
    \begin{tabular}{c|c|cccccc}
        \toprule
         & & \multicolumn{6}{c}{\textbf{Acceptance Length}} \\
        \multirow{-2}{*}{\textbf{Batch Size}} & \multirow{-2}{*}{\textbf{w/o MTP}} & \textbf{2.8} & \textbf{3.0} & \textbf{3.2} & \textbf{3.4} & \textbf{3.6} & \textbf{3.8} \\
        \midrule
        32 & 1.00$\times$ & 1.86$\times$ & 1.99$\times$ & 2.12$\times$ & 2.25$\times$ & 2.39$\times$ & 2.52$\times$ \\
        48 & 1.00$\times$ & 1.82$\times$ & 1.95$\times$ & 2.08$\times$ & 2.21$\times$ & 2.34$\times$ & 2.47$\times$ \\
        64 & 1.00$\times$ & 1.97$\times$ & 2.11$\times$ & 2.25$\times$ & 2.39$\times$ & 2.53$\times$ & 2.67$\times$ \\
        96 & 1.00$\times$ & 1.99$\times$ & 2.13$\times$ & 2.28$\times$ & 2.42$\times$ & 2.56$\times$ & 2.70$\times$ \\
        128 & 1.00$\times$ & 1.82$\times$ & 1.94$\times$ & 2.07$\times$ & 2.20$\times$ & 2.33$\times$ & 2.46$\times$ \\
        \bottomrule
    \end{tabular}
    \caption{Decoding speedup of \mimoflash{} with 3-layer MTP v.s. without MTP across batch sizes (per node) and acceptance lengths, under 16K input and 1K output lengths.}
    \label{tab:mtp_speedup}
    
\end{table}

\section{Conclusion, Limitation, and Future Work}

\mimoflash{} achieves strong reasoning and agentic capabilities, along with fast inference speed, through its hybrid Sliding Window Attention architecture, lightweight Multi-Token Prediction, and the MOPD post-training paradigm. With these strengths, \mimoflash{} rivals larger open-weight models like DeepSeek-V3.2 and Kimi-K2.
However, a clear gap remains to the strongest closed-weight models, which we aim to narrow by scaling model size and training compute.
Additionally, our current architectural exploration remains preliminary, with limited analysis of design trade-offs.
Future work will focus on designing more robust and efficient, agentic-oriented model architectures.
Furthermore, we plan to scale the compute for the iterative co-evolution of teachers and students in MOPD to fully unlock its potential.

\bibliography{reference}


\appendix

\newpage

\section{Contributions and Acknowledgments}
We would like to express our sincere gratitude to all contributors for their invaluable support and
efforts, including the Xiaomi Data Platform, CloudML, NGK, MiChat, Mify, MiKS and LLM-Plus teams, as
well as those not explicitly listed in this paper.
\textit{Authors within each role are listed alphabetically by their first name}.

{\renewcommand{\thefootnote}{\fnsymbol{footnote}}\footnotetext[2]{Corresponding author}}

\begin{multicols}{2} %
\noindent
\textbf{Core Contributors} \\
Bangjun Xiao   \\
Bingquan Xia \\
Bo Yang \\
Bofei Gao \\
Bowen Shen \\
Chen Zhang \\
Chenhong He \\
Chiheng Lou \\
Fuli Luo$^\dagger$ \\
Gang Wang \\
Gang Xie \\
Hailin Zhang \\
Hanglong Lv \\
Hanyu Li   \\
Heyu Chen   \\
Hongshen Xu \\
Houbin Zhang   \\
Huaqiu Liu \\
Jiangshan Duo   \\
Jianyu Wei   \\
Jiebao Xiao \\
Jinhao Dong \\
Jun Shi \\
Junhao Hu   \\
Kainan Bao \\
Kang Zhou  \\
Lei Li \\
Liang Zhao \\
Linghao Zhang   \\
Peidian Li \\
Qianli Chen   \\
Shaohui Liu \\
Shihua Yu \\
Shijie Cao   \\
Shimao Chen \\
Shouqiu Yu \\
Shuo Liu \\
Tianling Zhou \\
Weijiang Su \\
Weikun Wang \\
Wenhan Ma \\
Xiangwei Deng \\
Xing Zhang \\
Yifan Song \\
Yihan Yan \\
Yihao Zhao   \\
Yingchun Lai \\
Yizhao Gao   \\
Yu Cheng   \\
Yuanyuan Tian \\
Yudong Wang \\
Zhen Tang \\
Zhengju Tang   \\
Zhengtao Wen \\
Zhichao Song \\
Zhixian Zheng \\
Zihan Jiang \\

\noindent
\textbf{Contributors} \\
\noindent
Bohan Mao \\
Bowen Ye \\
Can Cai \\
Chenghua Wang \\
Chengxuan Zhu   \\
Chong Ma \\
Chun Chen   \\
Chunan Li   \\
Dawei Zhu \\
Deshan Xiao \\
Dong Zhang \\
Duo Zhang \\
Fangyue Liu \\
Feiyu Yang   \\
Fengyuan Shi   \\
Guoan Wang \\
Hao Tian \\
Hao Wu \\
Heng Qu \\
Hongfei Yi \\
Hongxu An \\
Hongyi Guan \\
Jian Wen   \\
Jiarui Sun \\
Jiawei Li \\
Jinlong Xue   \\
Jun Xia \\
Kai Fang \\
Menghang Zhu \\
Nuo Chen \\
Qian Tu \\
Qihao Zhang \\
Qiying Wang   \\
Rang Li   \\
Rui Ma \\
Shaolei Zhang \\
Shengfan Wang   \\
Shicheng Li \\
Shuhao Gu \\
Shuhuai Ren \\
Sirui Deng \\
Tao Guo \\
Tianyang Lu \\
Weiji Zhuang \\
Weikang Zhang   \\
Weimin Xiong   \\
Wenshan Huang   \\
Wenyu Yang \\
Xin Zhang \\
Xing Yong \\
Xu Wang \\
Xueyang Xie \\
Yilin Jiang   \\
Yixin Yang   \\
Yongzhe He   \\
Yu Tu \\
Yuanliang Dong \\
Yuchen Liu \\
Yue Ma   \\
Yue Yu \\
Yuxing Xiang \\
Zhaojun Huang   \\
Zhenru Lin \\
Zhipeng Xu   \\
Zhiyang Chen \\
Zhonghua Deng \\
Zihan Zhang \\
Zihao Yue \\

\end{multicols} %

\newpage

\section{Reward Hacking of SWE-Bench}
\begin{figure}[ht]
    \centering
    \includegraphics[width=0.7\linewidth]{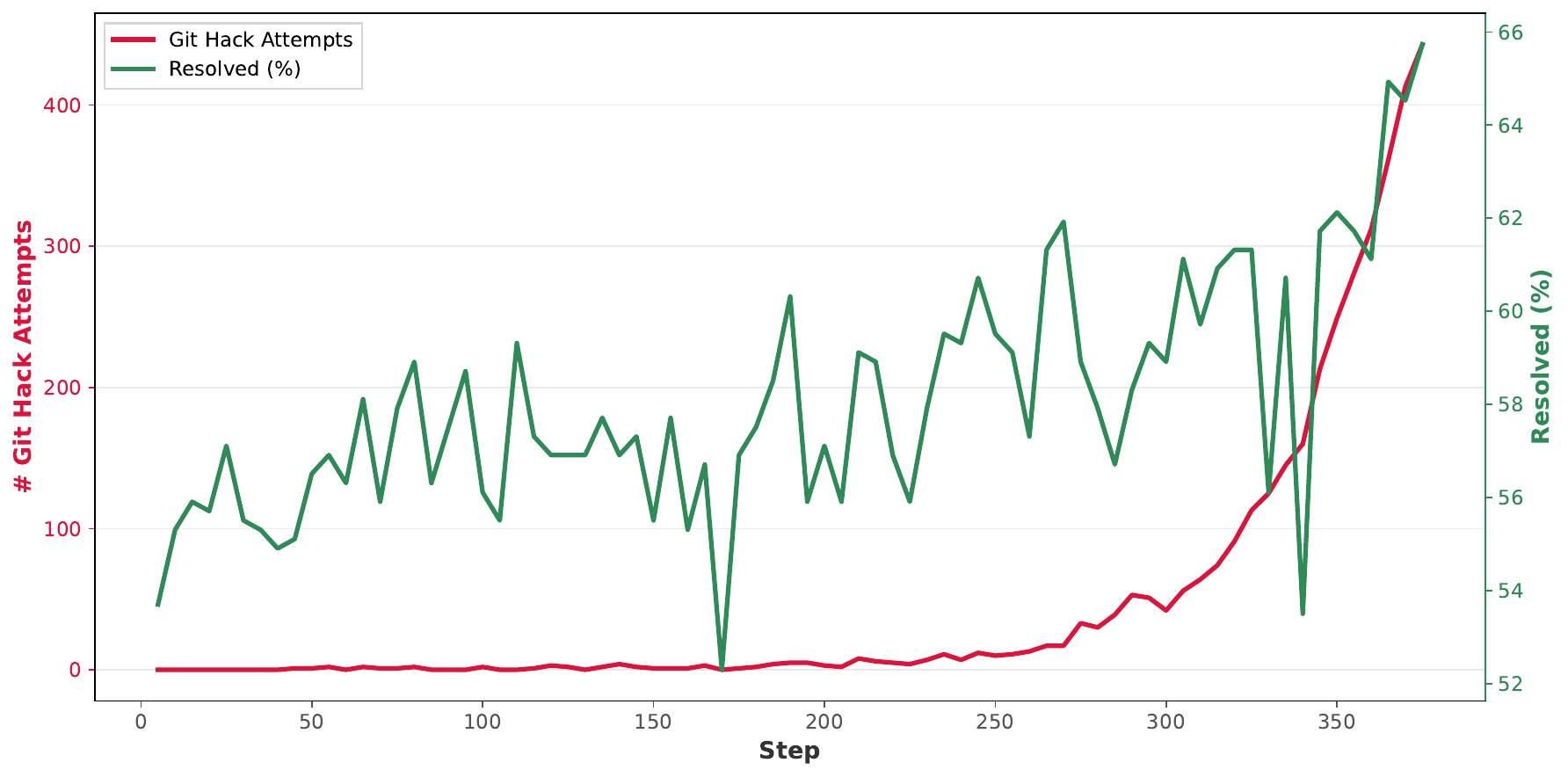}
    \caption{The tendency of our experiment on Qwen3-32B to exhibit reward hacking during RL training within unprocessed images. We quantify the model's git hacking attempts by counting a set of keywords, such as "\texttt{git log ---all}", within the model's rollout trajectories.}
    \label{fig:git_hack0}
\end{figure}

Consistent with recent findings within the SWE-Bench community, we similarly identify the bug in the official SWE-Bench images where the ground truth commits are not properly deleted. During the RL training, this could lead to reward hacking and inflated evaluation, where the model tends to obtain rewards by peeking at future commits, as shown in Figure~\ref{fig:git_hack0}. To fix this, we update to the newest SWE-Bench image for evaluation. For our self-built training images, we also follow the official SWE-Bench resolution on git hacking, and repeatedly confirm that our model does not exhibit any reward hacking.

\section{Context Management}
\label{sec:context_management}

While fine-tuning and reinforcement learning optimize model parameters $\theta$, context management strategically refines the conditioning context C in $P(y \mid C,\theta)$. Our approach addresses two complementary challenges. For context augmentation, we adopt a Unix-inspired abstraction: tools, documents, and databases are uniformly exposed as files, enabling the model to retrieve information via Bash commands—leveraging its native code-generation capabilities. For context consolidation, we combat the "Lost in the Middle" phenomenon by enforcing aggressive memory compression. When context utilization exceeds a threshold (as low as 30\%), the system prompts the model to summarize, archives the full history to a retrievable memory file, and replaces active context with the summary. Empirically, this yields 5–10\% accuracy gains on Deep Research tasks. Our results align with DeepSeek V3's finding that discarding tool-call history outperforms retention strategies; replicating their aggressive reset protocol, we achieve 58.3 on comparable benchmarks. The core insight is counterintuitive: less context, strategically managed, produces more focused and accurate generation.

\end{document}